\documentclass[sigconf]{acmart}
\AtBeginDocument{%
  \providecommand\BibTeX{{%
    \normalfont B\kern-0.5em{\scshape i\kern-0.25em b}\kern-0.8em\TeX}}}

\setcopyright{acmcopyright}
\copyrightyear{2023}
\acmYear{2023}
\acmDOI{XXXXXXX.XXXXXXX}

\acmConference[ICONS'23]{International Conference on Neuromorphic Systems}{August 1--3, 2023}{Santa Fe, NM}
\acmPrice{15.00}
\acmISBN{978-1-4503-XXXX-X/18/06}

\usepackage{hyperref}
\usepackage{ragged2e}
\usepackage{array}
\usepackage{multirow}

\begin{document}

\title{SuperNeuro: A Fast and Scalable Simulator for Neuromorphic Computing}

\author{Prasanna Date}
\email{datepa@ornl.gov}
\orcid{0000-0002-1664-069X}
\affiliation{%
  \institution{Oak Ridge National Laboratory}
  \city{Oak Ridge}
  \state{Tennessee}
  \country{USA}
}

\author{Chathika Gunaratne}
\email{gunaratnecs@ornl.gov}
\orcid{0000-0002-2508-8745}
\affiliation{%
  \institution{Oak Ridge National Laboratory}
  \city{Oak Ridge}
  \state{Tennessee}
  \country{USA}
}

\author{Shruti Kulkarni}
\email{kulkarnisr@ornl.gov}
\orcid{0000-0001-6894-9851}
\affiliation{%
  \institution{Oak Ridge National Laboratory}
  \city{Oak Ridge}
  \state{Tennessee}
  \country{USA}
}

\author{Robert Patton}
\email{pattonrm@ornl.gov}
\orcid{0000-0002-8101-0571}
\affiliation{%
  \institution{Oak Ridge National Laboratory}
  \city{Oak Ridge}
  \state{Tennessee}
  \country{USA}
}

\author{Mark Coletti}
\email{colettima@ornl.gov}
\affiliation{%
  \institution{Oak Ridge National Laboratory}
  \city{Oak Ridge}
  \state{Tennessee}
  \country{USA}
}

\author{Thomas Potok}
\email{potokte@ornl.gov}
\orcid{0000-0001-6687-3435}
\affiliation{%
  \institution{Oak Ridge National Laboratory}
  \city{Oak Ridge}
  \state{Tennessee}
  \country{USA}
}



\renewcommand{\shortauthors}{Trovato and Tobin, et al.}

\begin{abstract}
In many neuromorphic workflows, simulators play a vital role for important tasks such as training spiking neural networks (SNNs), running neuroscience simulations, and designing, implementing and testing neuromorphic algorithms. Currently available simulators are catered to either neuroscience workflows (such as NEST and Brian2) or deep learning workflows (such as BindsNET). While the neuroscience-based simulators are slow and not very scalable, the deep learning-based simulators do not support certain functionalities such as synaptic delay that are typical of neuromorphic workloads. In this paper, we address this gap in the literature and present SuperNeuro, which is a fast and scalable simulator for neuromorphic computing, capable of both homogeneous and heterogeneous simulations as well as GPU acceleration. We also present preliminary results comparing SuperNeuro to widely used neuromorphic simulators such as NEST, Brian2 and BindsNET in terms of computation times. We demonstrate that SuperNeuro can be approximately 10--300 times faster than some of the other simulators for small sparse networks. On large sparse and large dense networks, SuperNeuro can be approximately 2.2 and 3.4 times faster than the other simulators respectively.
\end{abstract}






\maketitle

\section{Introduction}
\label{sec:intro}

Neuromorphic computing is a promising computing paradigm for low-power applications \cite{schuman2022opportunities}, which span neuroscience simulations, autonomous vehicles, anomaly detection, graph algorithms, epidemiological modeling, scientific applications etc. \cite{patton2021neuromorphic, hamilton2020modeling, cong2022semi,date2018efficient,hamilton2020spike,markram2006blue}.
If used in a machine learning setting, training spiking neural networks (SNNs) is a critical component of the neuromorphic workflow---this is usually done on CPUs or GPUs \cite{bhuiyan2010acceleration}.
If used in a non-machine learning setting, it is important to design, implement and test novel neuromorphic algorithms efficiently and rapidly \cite{aimone2022review}.
Both the above use cases benefit from an efficient simulator. 

Unlike CPUs and GPUs, neuromorphic hardware is not readily available off-the-shelf \cite{kulkarni2021benchmarking}. 
It is typically available at research groups scattered throughout the globe and only as research-grade prototype devices such as Intel Loihi, IBM TrueNorth and SpiNNaker \cite{davies2018loihi,debole2019truenorth,furber2014spinnaker}. 
As a result, it is difficult to work with the neuromorphic hardware directly, especially for researchers who design novel neuromorphic algorithms or train SNNs for their applications. 
To mitigate this lack of availability of neuromorphic hardware, an efficient neuromorphic computing simulator is indispensable.

Neuromorphic algorithms inform the design of the neuromorphic hardware in many ways \cite{schuman2022opportunities}. 
For instance, some neuromorphic algorithms require dense connectivity, whereas others require sparse connectivity. 
Some algorithms require floating point synaptic weights, whereas others require binary (0, 1) weights. 
To enable this co-design between neuromorphic algorithms and hardware, simulators play a vital role. 
They establish the requirements for the neuromorphic algorithms, making it possible to realize these requirements on the hardware.

The current neuromorphic computing simulators are primarily focused on either neuroscience workloads, or deep learning workloads \cite{kulkarni2021benchmarking}. 
The neuroscience-based simulators such as NEST (based on discrete event simulation), and Brian2 (based on solving system of ordinary differential equations) are slow for SNN-based machine learning and general-purpose computing applications \cite{Gewaltig:NEST,stimberg2019brian,kulkarni2021benchmarking}. 
On the other hand, the deep learning-based simulators such as snntorch and BindsNET do not provide all the features---for instance, synaptic delays---needed for neuroscience and general-purpose neuromorphic computing applications \cite{eshraghian2021training, hazan2018bindsnet}. 

To this extent, we present SuperNeuro, a fast and scalable Python-based simulator for neuromorphic computing with GPU acceleration capability. 
SuperNeuro models the neuromorphic simulation problem using two different approaches: a matrix computation-based approach (MAT) and an agent-based modeling (ABM) approach. 
To the best of our knowledge, the neuromorphic simulation problem has not been modeled using these approaches in the literature.
SuperNeuro provides a development framework for accelerating neuromorphic simulations with the flexibility to define custom neuron and synapse models, while utilizing computationally efficient algorithms and hardware for highly scalable simulations. 
SuperNeuro provides the AI practitioner the capability of studying and optimizing SNNs of large-scale regardless of the current availability of neuromorphic hardware. 
It also allows the practitioner to implement general-purpose computing algorithms such as graph algorithms, non-neural network-based machine learning algorithms, and design neuromorphic primitives for data encoding and data structures.

\section{SuperNeuro}
\label{sec:superneuro}

\begin{table}[t!]
    \caption{SuperNeuro Modes: Matrix Computation (MAT) Mode and Agent-Based Modeling (ABM) Mode}
    \centering
    \begin{tabular}{m{0.2\textwidth} m{0.22\textwidth}}
        \noalign{\smallskip} \hline \noalign{\smallskip}
        MAT Mode & ABM Mode  \\
        \noalign{\smallskip} \hline \noalign{\smallskip}
        Homogeneous simulations & Heterogeneous simulations\\
        \noalign{\smallskip}
        Built-in STDP learning & No built-in learning \\
        \noalign{\smallskip}
        CPU execution & GPU acceleration  \\
        \noalign{\smallskip}
        Fast and scalable  & Custom neurons and synapses \\
        \noalign{\smallskip} \hline \noalign{\smallskip} 
    \end{tabular}
    \label{tab:modes}
\end{table}

SuperNeuro is a fast and scalable simulator for neuromorphic computing. It lends itself to different types of workloads, which include neuroscience workloads, deep learning (based on SNNs) workloads and general-purpose computing workloads. 
SuperNeuro features two modes that model the neuromorphic simulations in two different ways: a matrix computation-based mode (MAT) and an agent-based modeling (ABM) mode.
Table \ref{tab:modes} shows the features of the two SuperNeuro modes. 
A brief description of these modes is provided in the following subsections.
The goal of this short paper is to highlight preliminary computational results for SuperNeuro in comparison to other widely used simulators---this is covered in Section \ref{sec:results}.
While we go over the inner workings of MAT and ABM modes briefly, we do not explain the mathematical and algorithmic details in this paper---these details remain out of scope for this paper and will be explained in detail in upcoming papers.
The SuperNeuro code is open-source and available on GitHub at \url{https://github.com/ORNL/superneuro}.


\subsection{The MAT Mode}
\label{sub:mat}

The MAT mode in SuperNeuro models neuromorphic simulations using matrices and vectors.
It supports homogeneous simulations, i.e., simulations where all neurons and all synapses are of the same type.
All neurons in the MAT mode are leaky integrate and fire (LIF) neurons.
Each neuron has four parameters: threshold, leak, reset state and refractory period.
The leak used in the MAT mode is of constant type. 
Specifically, at each time step, a constant value is subtracted from (or added to) the internal state of each neuron in order to bring the internal state closer to the reset state.
So, if the internal state is greater than the reset state (but less than the threshold) we subtract the leak; else, we add the leak.
The internal states of all neurons at a given time step in the simulation are represented as a vector. 
The neuron thresholds, leaks, reset states, refractory periods, and spikes are also represented as vectors.

Each synapse in the MAT mode has two parameters: weight and delay.
The weights are represented in a square matrix such that the weight of the synapse going from neuron $i$ to neuron $j$ is captured at the element located at the $i^{\text{th}}$ row and $j^{\text{th}}$ column in the matrix.
The delay is computed by adding proxy neurons in the simulations such that each and every synapse in the simulation always has a delay of unity. 
For example, say that a synapse from neuron $i$ to neuron $j$ needs to have a delay of three.
We implement this functionality by adding two proxy neurons $k_1$ and $k_2$, both having a threshold of zero and infinite (very high) leak.
Next, we connect neuron $i$ to neuron $k_1$, neuron $k_1$ to neuron $k_2$, and neuron $k_2$ to neuron $j$, such that each of these synapses have a weight and delay of unity.
The effective delay going from neuron $i$ to neuron $j$ is thus three as required.
While this implementation of synaptic delay sidesteps any temporal computational overheads, it does introduce spatial overheads by adding proxy neurons. 
As a result, this approach could be inefficient for neuromorphic simulations that contain significant amount of delays on the synapses.
In general, all neuron and synapse operations are represented as matrix or vector operations in the MAT mode and thus, can be easily parallelized. 

To compute the internal states of all neurons at the current time step, we first start with the internal state vector from the previous time step and apply the constant leak to it.
Second, the vector of spikes from the previous time step is multiplied by the weight matrix and then added to the internal state vector from the previous time step.
Third, if there are any external spikes at the current time step, they are added to the internal state vector of the current time step. 
Next, the spike vector at current time step is computed by comparing the current internal state vector to the vector of neuron thresholds. 
Lastly, for neurons in their refractory periods, the spikes are zeroed out.
All the above operations are implemented in numpy, which is a highly efficient, CPU-based, numerical computation library in Python. 

When the problem is formulated in this fashion, it allows for both speed and scalability for homogeneous simulations. 
The MAT mode also has a built-in Spike-Time Dependent Plasticity (STDP) learning mechanism, which can be used for training SNNs.
The STDP mechanism is also implemented using matrix and vector operations.
As shown in Table \ref{tab:modes}, SuperNeuro supports homogeneous simulations in the MAT mode. 
At present, the MAT mode only supports CPU execution but GPU acceleration capability will be provided in the future. 
The MAT mode should be used in cases where speed and scalability are important.

\subsection{The ABM Mode}
\label{sub:abm}

Agent-based modeling (ABM) is a widely used modeling and simulation technique for investigating complex adaptive systems. 
In SuperNeuro's ABM mode, we take an organic perspective to SNN design, simulating each neuron as an individual agent. 
The simulation is clock-driven, with step functions defining the actions taken by agents at each time step. 
In the ABM mode, the neurons and synapses inherit the autonomy of the agent modeling paradigm allowing for heterogeneous simulations, i.e., the neurons and synapses in the simulation can be of different types, each supporting a unique spiking mechanism. 
The unique features available in the ABM mode are shown in Table \ref{tab:modes}.
This heterogeneous simulation feature is useful for exploring neuron and synapse mechanisms that have not been realized on the hardware yet, for example, stochastic neurons, stochastic synapses etc.

The ABM mode is implemented using SAGESim, a GPU-capable agent-based modeling framework developed at the Oak Ridge National Laboratory. 
While using GPUs, SAGESim works by assigning a GPU thread to each agent. SAGESim supports the execution of multi-breed simulation and multiple step functions, ordered by priority for each breed.
As mentioned before, each neuron in the ABM mode is an agent within the SAGESim simulation platform.
Each neuron agent is provided with two step functions: the neuron step function and synapse step function. The neuron step function aggregates external spikes at the current timestep with the current internal state of the neuron and subtracts the leak. If the new internal state is greater than the neuronal threshold, the neuron spikes and the delay registers of outgoing synapses are updated accordingly. 
In the synaptic step function, the synaptic delay registers are updates and the final elements of the delay registers are used to update the internal states of the post-neurons.

The agent-based form is well suited for GPU acceleration and greatly improves speed and scalability. 
The ABM mode supports both CPU and GPU execution; GPU device must be NVIDIA compute capability 6 or higher.
It currently supports the leaky integrate and fire mechanism for the neuron.
Furthermore, the neuron and synapse parameters supported include threshold,  leak,  refractory period, axonal delay, synaptic weights, and synaptic delays.
The notion of leak supported in the ABM mode is the same as the one in the MAT mode.
The ABM mode does not support any learning mechanism at present, but will support STDP learning in the future.

\section{Results}
\label{sec:results}

\begin{table*}[h]
    \centering
    \begin{tabular}{l  c  c  c  c  c | c  c  c  c | c  c  c  c}\toprule
\multicolumn{2}{r}{Number of Neurons} & \multicolumn{4}{c}{100} & \multicolumn{4}{c}{1000} & \multicolumn{4}{c}{10000}\\\midrule
\multicolumn{2}{r}{Connection Probability} & 0.25  & 0.50 & 0.75 & 1.0 & 0.25 & 	0.50 & 0.75	& 1.0 & 0.25 & 0.50	& 0.75 & 1.0\\\midrule
\multirow{2}{*}{SuperNeuro} & MAT & \textbf{0.04}  & \textbf{0.04}	 & \textbf{0.05}  & \textbf{0.05}	  & \textbf{0.36}    & \textbf{0.52}	  & \textbf{0.68}    & \textbf{0.80}    & \textbf{30.37}   & \textbf{40.34}  & \textbf{55.38}  & \textbf{71.62}\\
	                      & ABM & 1.10  & 1.16   & 1.18  & 1.26   & 4.11    & 5.59    & 7.22    & 9.08    & 166.72  & 325.30 & 488.83 & 641.36\\
\multicolumn{2}{l}{NEST}          & 0.41  & 0.65   & 0.91  & 1.23   & 28.05   & 61.42   & 81.54   & 112.82  & 3242.83 & > 1h   & > 1h   & > 1h\\
\multicolumn{2}{l}{Brian2}        & 6.63  & 12.43  & 18.36 & 24.23  & 612.43  & 1249.31 & 1892.19 & 2529.77 & > 1h    & > 1h   & > 1h   & > 1h\\
\multicolumn{2}{l}{BindsNET}      & 2.44  & 2.34   & 2.46  & 2.38   & 3.45    & 3.67    & 4.19    & 4.64    & 63.73   & 117.78 & 181.04 & 230.07\\\bottomrule
    \end{tabular}
    \caption{Total execution times in seconds for SuperNeuro vs other state-of-the-art simulators for different configurations of network sizes, given by number of neurons, and network sparsities, given by synapse connection probabilities.}
    \label{tab:execution-times}
\end{table*}

We compare the performance of both SuperNeuro modes against three widely used neuromorphic simulators: NEST, Brian2, and BindsNET. 
We generated random networks using a graph and network algorithms library in Python called networkx.
Using the Erdős-Rényi graph generation algorithm in networkx, we generated random graphs such that each node of the graph would correspond to a neuron and each edge of the graph would correspond to a synapse in our neuromorphic simulations. 
Networks of 100, 1000, and 10,000 neurons were generated as networks of these sizes can be efficiently run on a desktop workstation. 
For each size of the network we varied the sparsity by changing the connection probabilities of the synapses. 
We chose the following values for the synapse connection probabilities: 0.25, 0.5, 0.75, and 1. 
With three values for number of neurons and four values for the synapse connection probabilities, we had 12 network configurations. 
Each of these network configurations was used to initialize the neurons and synapses within each of the five simulators. 
Each simulation was run for 1000 time steps and 3 input neurons were chosen at random. 
Each input neuron was externally spiked at every 10 time steps. 
All neuron thresholds were set to 1, reset states were set to 0, refractory periods were set to 0, and the axonal delays (if applicable) were set to 0.
All synaptic weights and delays were set to 1. 

Table \ref{tab:execution-times} displays the total execution times in seconds for all the simulator runs. 
The runs that took longer than an hour were terminated and are depicted with \textit{> 1h}. 
All ABM and BindsNET runs leveraged GPU acceleration.
The MAT mode in SuperNeuro was seen to be the fastest across all network configurations.
For the small-sized jobs (100 neurons), MAT obtained a speed up of 9x for the 0.25 connection probability over the next best simulator, which in this case, was NEST.
With increasing connection probabilities, MAT obtained increasing speedups of 15x, 20x and 27x over NEST.
As compared to some of the slower simulators such as Brian2, MAT obtained speedups of 150x, 290x, 406x and 530x for the four sparsity configurations.
The total execution times for ABM were comparable to NEST for the configuration with 100 neurons and 1.0 connection probability.
For this configuration, ABM obtained a speedup of 19x and 2x over Brian2 and BindsNET.

For the medium-sized networks (1000 neurons), BindsNET performed better than all other simulators except MAT, which obtained speedups of 10x, 7x, 6x and 6x over BindsNET for the four sparsities.
MAT also obtained speedups ranging from 1,695--3,174x over Brian2 for the different sparsities.
ABM was 7x, 11x, 11x and 12x faster than NEST for the four sparsities.
As compared to Brian2, ABM's speedup ranged from 149--279x for the four sparsities.
For the large-sized networks (10,000 neurons), BindsNET was one again faster than other simulators except MAT.
In this case, MAT obtained a speedup of 2-3x over BindsNET for the four different configurations.
For the 10,000 neuron and 0.25 connection probability configuration, MAT was 107x faster than NEST and possibly even faster on the higher connection probabilities.
The speedups for NEST and Brian2 for the 10,000 neuron configurations are not determined because those jobs took more than an hour and were terminated.
For these larger configurations, ABM took roughly a third of the time as compared to BindsNET.
For the 0.25 sparsity configuration, ABM performed 19x faster than NEST.
It also performed significantly faster than all other runs for NEST and Brian2.
Overall, on smaller-sized networks, MAT was the fastest, followed by NEST, ABM, BindsNET and Brian2.
On medium and large-sized networks, MAT was the fastest, followed by BindsNET, ABM, NEST and Brian2.

\section{Discussion}
\label{sec:discussion}

An efficient neuromorphic simulator allows us to work with customizable neuron and synapse mechanisms that go beyond the typical deferential equation specifications found in traditional neuroscience. This enables engineering and computer science focused design of SNNs for neuromorphic devices in an accelerated manner. 
Unlike existing neuromorphic simulators, SuperNeuro provides the capability for the practitioner to include heterogenous neuron and synapse spiking mechanisms within the same SNN. 

SuperNeuro can simulate larger spiking neural networks in less time with higher computational efficiency than any other neuromorphic simulator available today.
It can even simulate networks of the scale of some living organisms, comparable to those studied in neuroscience. 
For instance, we have simulated networks with 100,000 neurons (lobster-sized brain) with all-to-all connectivity on a desktop in approximately 5 minutes using the MAT mode. 
The current simulators, such as NEST and Brian2, take more than 1 hour for such tasks. 
By leveraging high performance computing resources at the Oak Ridge National Laboratory, SuperNeuro could potentially simulate networks with a few million neurons, such as those found in bees and lizards.

SNNs have been used to realize various cognitive and machine learning algorithms such as control, reinforcement learning, classification, decision trees, and regression. For the rapid development and benchmarking of these algorithms in the neuromorphic domain, the availability of a performant simulator is crucial. With the use of a performant simulator such as SuperNeuro, we can train SNNs for deployment on edge platforms for applications such as autonomous vehicles, industrial robotics, autonomous drones, high energy physics etc.
This enables AI practitioners to rapidly develop and prototype new SNN architectures significantly faster, and in turn, enable the co-design of neuromorphic hardware.

\section{Conclusion}
\label{sec:conclusion}

Currently, neuromorphic computing suffers from lack of fast, highly scalable neuromorphic, and flexible simulators for designing and training spiking neural networks. 
SuperNeuro provides AI practioners with a neuromorphic simulator in Python that is both fast and scalable and also provides the option of simulating the user’s own spiking mechanisms. 
SuperNeuro is capable of taking advantage of GPU acceleration and is able to provide superior performance as compared to existing simulation platforms. SuperNeuro can easily integrate with learning and optimization tools for spiking neural network optimization. This opens many possibilities for the successful co-design of neuromorphic circuits enabling intelligent edge computing device design, while at the same time facilitates large scale AI experimentation on accelerated computing infrastructure. 

\begin{acks}
This material is based in part upon work supported by  the U.S. Department of Energy, Office of Science, Office of Advanced Scientific Computing Research, under award number DE-SC0022566. 
\end{acks}

\bibliographystyle{acm}
\bibliography{references}

\end{document}